\pdfoutput=1

\documentclass[11pt]{article}

\usepackage[preprint]{acl}

\usepackage{times}
\usepackage{latexsym}
\usepackage{amsfonts}
\usepackage{graphicx}
\usepackage{svg}
\usepackage{multirow}
\usepackage{bm}
\usepackage{calligra}
\usepackage{booktabs}
\usepackage[T1]{fontenc}

\usepackage[utf8]{inputenc}

\usepackage{microtype}

\newcommand{\green}[1]{\textcolor{green}{#1}}
\newcommand{\cyan}[1]{\textcolor{cyan}{#1}}
\newcommand{\orange}[1]{\textcolor{orange}{#1}}

\title{QQQ: Quality Quattuor-Bit Quantization for Large Language Models}

\author{Ying Zhang$^{1}$, Peng Zhang$^{1}$, Mincong Huang$^{1}$, Jingyang Xiang$^{1,2}$, Yujie Wang$^{1}$\\
\bf Chao Wang$^{1}$, Yineng Zhang$^{1}$, Lei Yu$^{1}$, Chuan Liu, Wei Lin \\
\textsuperscript{\rm1}Meituan \textsuperscript{\rm2}Zhejiang University\\
{$\;\:$\texttt{\href{mailto:yingzhang1998@buaa.edu.cn}{yingzhang1998}@buaa.edu.cn}}}

\begin{document}
\maketitle
\begin{abstract}
Quantization is a proven effective method for compressing large language models. Although popular techniques like W8A8 and W4A16 effectively maintain model performance, they often fail to concurrently speed up the prefill and decoding stages of inference. W4A8 is a promising strategy to accelerate both of them while usually leads to a significant performance degradation.
To address these issues, we present QQQ, a \textbf{Q}uality \textbf{Q}uattuor-bit \textbf{Q}uantization method with 4-bit weights and 8-bit activations. QQQ employs adaptive smoothing and Hessian-based compensation, significantly enhancing the performance of quantized models without extensive training.
Furthermore, we meticulously engineer W4A8 GEMM kernels to increase inference speed. 
Our specialized per-channel W4A8 GEMM and per-group W4A8 GEMM achieve impressive speed increases of 3.67$\times$ and 3.29 $\times$ over FP16 GEMM.
Our extensive experiments show that QQQ achieves performance on par with existing state-of-the-art LLM quantization methods while significantly accelerating inference, achieving speed boosts up to 2.24 $\times$, 2.10$\times$, and 1.25$\times$ compared to FP16, W8A8, and W4A16, respectively. Code is available at \url{https://github.com/HandH1998/QQQ}.
\end{abstract}

\section{Introduction}
In recent years, large language models (LLMs) have been widely applied in various tasks and shown excellent performance, such as OPT \citep{zhang2022opt} and LLaMA series \cite{touvron2023llama, touvron2023llama2}.
The LLMs usually have large number of parameters and long inference time, making it inapplicable to resource-limited devices and real-time scenarios.
Therefore, it is crucial to reduce LLMs' storage and computation overhead while retaining their performance. 

Quantization is a crucial technique for reducing the memory and computational requirements of LLMs. It generally involves two main strategies: weight-activation and weight-only quantization. The W8A8 weight-activation quantization method, as demonstrated by SmoothQuant \citep{xiao2023smoothquant}, compresses both weights and activations to 8-bit, leveraging specialized hardware like INT8 Tensor Cores to increase computational throughput. This method excels in enhancing the compute-bound prefill phase of LLM inference.
Weight-only quantization, such as the W4A16 method seen in GPTQ \citep{frantar2022gptq} and AWQ \citep{lin2023awq}, reduces the weight precision to 4-bit while maintaining 16-bit activations. This approach effectively cuts memory bandwidth requirements and is particularly beneficial during the memory-bound decoding phase of LLM inference.
Although these methods successfully retain model performance, they struggle to simultaneously accelerate both the prefill and decoding stages of LLM inference. The W4A8 weight-activation quantization approach, which can expedite both inference phases by utilizing efficient hardware and reducing memory bandwidth, often leads to a significant decrease in model performance.
Therefore, the pursuit of a viable W4A8 quantization method that balances performance with speed remains an active area of research.

In response to these challenges, we introduce QQQ, a quality W4A8 quantization approach that excels in both model performance and inference speed. QQQ employs an adaptive smoothing mechanism for activation channels with significant outliers, while leaving other channels intact. This targeted smoothing allows for more effective activation quantization. Following the smoothing process, QQQ utilizes a Hessian-based compensation technique to offset the loss from weight quantization. Through these innovative strategies, QQQ significantly improves the performance of quantized models.
To achieve fast inference speed, we meticulously design specialized W4A8 GEMM kernels. These kernels are specifically tailored for both per-channel and per-group weight quantization, drastically enhancing throughput across various batch sizes.
Our contributions compared to previous works are threefold:

\begin{itemize}
\item
We present QQQ, a quality W4A8 quantization method, that delivers performance on par with existing state-of-the-art LLM quantization methods.
\item 
We develop innovative W4A8 GEMM kernels tailored for both per-channel weight and per-group weight quantization. These specialized kernels significantly enhance speed, with per-channel W4A8 GEMM and per-group W4A8 GEMM achieving speedups of 3.67$\times$ and 3.29$\times$ over FP16 GEMM, respectively.
\item 
We integrate QQQ into vLLM \citep{kwon2023efficient}, achieving impressive speed boosts of 2.24$\times$, 2.10$\times$, and 1.25$\times$ over FP16, W8A8, and W4A16 Marlin \citep{frantar2024marlin} implementation, respectively.
\end{itemize}


\section{Related Works}\label{section:sec2}

\subsection{Large Language Models}
In recent times, large language models (LLMs) such as GPT-3 \citep{brown2020language}, GLM \citep{du2021glm}, OPT \citep{zhang2022opt}, and the LLaMA series \cite{touvron2023llama, touvron2023llama2} have revolutionized the field of natural language processing by delivering unprecedented performance enhancements across various tasks. Despite their impressive capabilities, LLMs come with a significant computational footprint due to their billions of parameters, which takes long inference time. This makes them less suitable for deployment on devices with limited resources or in applications requiring real-time responses. For example, the GPT-3 \citep{brown2020language} model, with its colossal count of 175 billion parameters, demands a staggering 350 GB of memory for loading its parameters in FP16 format. This necessitates the use of at least five A100-80G GPUs solely for inference purposes.
\subsection{Quantization}
Current strategies for quantizing LLMs fall into two main categories: weight-activation and weight-only quantization.

\textbf{Weight-activation quantization.} Weight-activation quantization compresses both weights and activations to lower bit representations such as W8A8, W4A8, W6A6, and W4A4. SmoothQuant \citep{xiao2023smoothquant}, LLM.int8() \citep{dettmers2208llm}, QUIK \citep{ashkboos2023towards} and Outlier Suppression+ \citep{wei2023outlier} all enhance quantized model performance by managing activation outliers. LLM.int8() \citep{dettmers2208llm} and QUIK \citep{ashkboos2023towards} employ mixed-precision decomposition, while SmoothQuant \citep{xiao2023smoothquant} and Outlier Suppression+ \citep{wei2023outlier} use channel-wise scaling. RPTQ \citep{yuan2023rptq} mitigates the impact of range differences between channels by rearranging the channels and quantizing them in clusters. OmniQuant \citep{shao2023omniquant} proposes learnable weight clipping and learnable equivalent transformation. Atom \citep{zhao2023atom} attains high accuracy by applying a novel mixed-precision and fine-grained quantization process.  QuaRot \citep{ashkboos2024quarot} removes outliers by rotating the inputs of the model using randomized Hadamard transformations.

\textbf{Weight-only quantization.} Weight-only quantization compresses model weights to lower bit precision (e.g., 4-bit or 2-bit) while keeping activations at higher precision (16-bit). {GPTQ \citep{frantar2022gptq} adopts second-order information to minimize precision loss.} SpQR \citep{dettmers2023spqr} and AWQ \citep{lin2023awq} prioritize the preservation of important weights to reduce quantization errors, with SpQR applying mixed-precision and AWQ optimizing per-channel scaling. QuIP \citep{chee2024quip} further quantizes weights to 2-bit based on the insight that model parameters should ideally be incoherent.

\section{Methodology}\label{section:sec4}
\begin{figure}[ht]
\centering
\includegraphics[scale=0.8]{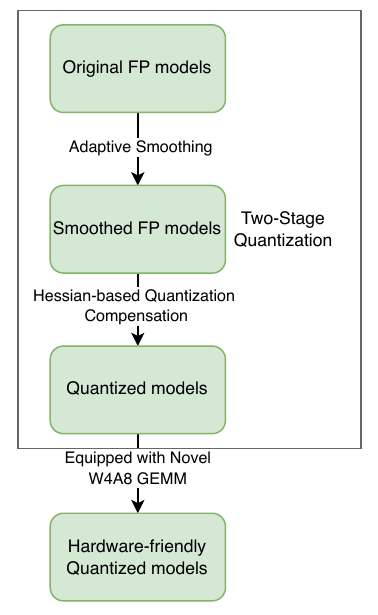}
\caption{Framework of QQQ.}
\label{fig:fig1}
\vspace{-0.1in}
\end{figure}

In this section, we detail our quantization method, referred to as QQQ. As depicted in Figure \ref{fig:fig1}, QQQ is a two-stage weight-activation quantization method.
It integrates adaptive smoothing and Hessian-based quantization compensation to enhance the performance of quantized models. QQQ adaptively smooths activation channels with significant outliers while leaving other channels intact, enabling better activation quantization.
After smoothing the activations, QQQ tackles the challenges of weight quantization by employing Hessian-based compensation, which effectively minimizes the loss incurred during the quantization process.
To expedite practical inference, QQQ incorporates highly efficient W4A8 GEMMs tailored for both per-channel and per-group quantization. The quantization process will be further elaborated in the following subsections.

\subsection{Adaptive Smoothing}
We employ channel-wise smoothing to refine problematic activation channels with outliers, enhancing their suitability for quantization. As indicated by research \citep{wei2023outlier}, smoothing only the activation channels with significant outliers while leaving other channels intact can enhance the performance of quantized models. Therefore, we further implement an adaptive smoothing strategy that identifies and selectively smooths only those channels that display significant outliers.

Specifically, we introduce an outlier threshold $\sigma$, which is determined through a grid search within the range $[0, \mbox{max}(\mbox{abs}(\mathbf{X}))]$. Activation channels that exhibit outliers exceeding this threshold $\sigma$ are then selectively smoothed:
\begin{equation}
    \mathcal{T}=\{\,t\,|\,\mbox{max}(\mbox{abs}(\mathbf{X}_{:, t}))<\sigma \},
    \label{eq:selct_channel}
\end{equation}
where $\mathcal{T}$ denotes the selected channel index for $\mathbf{X}$.

To optimize the smoothing parameters, we aim to reduce the squared error between the product of activations and weights both before and after quantization:
\begin{equation}
\arg\min_{\mathbf{s}}\|Q(\mathbf{X}_{:,\mathcal{T}} \oslash \mathbf{s})Q(\mathbf{W}_{\mathcal{T}, :} \odot \mathbf{s})- \mathbf{X}_{:,\mathcal{T}}\mathbf{W}_{\mathcal{T}, :}\|_2^2,
\end{equation}
where $Q$ denotes the quantization function;
$\mathbf{W}$ denotes the unquantized weight matrix;
$\mathbf{X}$ denotes the unquantized activation matrix;
$\mathbf{s}$ denotes the smoothing parameter;
$\oslash$ and $\odot$ are element-wise division and multiplication, respectively.



\subsection{Hessian-based Quantization Compensation}
After transferring the quantization challenge from activations to weights through adaptive smoothing, it becomes crucial to enhance the quantization technique for weights. We adopt a layer-wise quantization framework GPTQ \citep{frantar2022gptq}, which advocates for a precise and efficient Hessian-based quantization compensation technique. This method involves an application of two key formulas, Eq.(\ref{eq:eq2}) and Eq.(\ref{eq:eq3}), which are instrumental in the iterative quantization of weights. Specifically, this algorithm adjusts the subset $F$ of full-precision weights by an increment $\delta_F$, compensating for the quantization error induced by the quantized weights $Q(\textbf{W}_i)$.
\begin{equation}
    \mathbf{W}_i = \mathop{\mbox{argmin}}\limits_{\mathbf{W}_i}\frac{(Q(\textbf{W}_i) - \textbf{W}_i)^2}{[\mathbf{H}_{F}^{-1}]_{ii}} \label{eq:eq2}
\end{equation}
\begin{equation}
    \bm{\delta}_F = - \frac{\textbf{W}_i - Q(\textbf{W}_i)}{[\mathbf{H}_{F}^{-1}]_{ii}} \cdot (\mathbf{H}_{F}^{-1})_{:,i} \label{eq:eq3}
\end{equation}
where $\textbf{W}_i$ denotes the $i$-row of weight matrix $\mathbf{W}$, $\mathbf{H}_{F} = 2\textbf{X}_F^\mathsf{T}\textbf{X}_F$ denotes the Hessian matrix.

\subsection{Novel W4A8 GEMM}
We carefully develop innovative W4A8 GEMMs tailored to optimize the inference speed. It's worth noting that our design exclusively supports symmetric quantization due to its superior hardware efficiency compared to asymmetric quantization.

The W4A8 GEMM design introduces two critical considerations beyond conventional GEMM implementations.
Firstly, as GPUs support GEMM operations only for identical data types, it's necessary to convert INT4 weights into INT8 to utilize the capabilities of the INT8 Tensor Cores.
Secondly, to avoid potential overflow during INT8 GEMM, the final output is typically stored in INT32 format, which subsequently requires a dequantization setp from INT32 to FP16.
%

%
To further reduce memory access, we integrate both the type conversion and dequantization process within the W4A8 GEMM routine.
We have different W4A8 GEMM configurations for per-channel weight quantization and per-group weight quantization, as they have different quantization schemes, which will be detailed in the following subsections.

\begin{figure}[ht]
\centering
\includegraphics[scale=0.65]{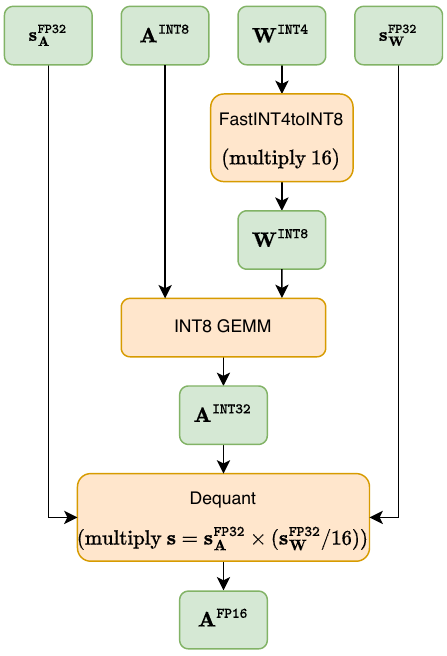}
\caption{The W4A8 GEMM for per-channel weight quantization. $\mathbf{A}$ denotes the activation matrix. $\mathbf{W}$ denotes the weight matrix. $\mathbf{s_A}$ denotes the per-token quantization scale of activations. $\mathbf{s_W}$ denotes the per-channel quantization scale of weights. The superscript represents the data type. The green sections indicate \green{data}, and the orange sections indicate \orange{operations}.}
\label{fig:fig3}
\vspace{-0.1in}
\end{figure}

\subsubsection{Per-channel Weight Quantization}
The specialized W4A8 GEMM process for per-channel weight quantization is depicted in Figure \ref{fig:fig3}. Our method involves an initial conversion of INT4 weights to INT8 before executing the \texttt{INT8 GEMM}, followed by a dequantization step that transforms the INT32 output of the \texttt{INT8 GEMM} into a FP16 result. To ensure high efficiency in the type conversion, we utilize the \texttt{FastINT4toINT8} \citep{li2023speed} technique. This process involves positioning the INT4 weight into the upper 4 bits of an INT8 by multiplying by 16, essentially performing a left shift by 4 bits. Once the GEMM computation is complete, we scale down the result by a factor of 16 to obtain the accurate output. This scaling operation is seamlessly integrated into the dequantization step, \texttt{Dequant}, by adjusting the weight quantization scale offline by the same factor. 

\begin{figure}[ht]
\centering
\includegraphics[scale=0.65]{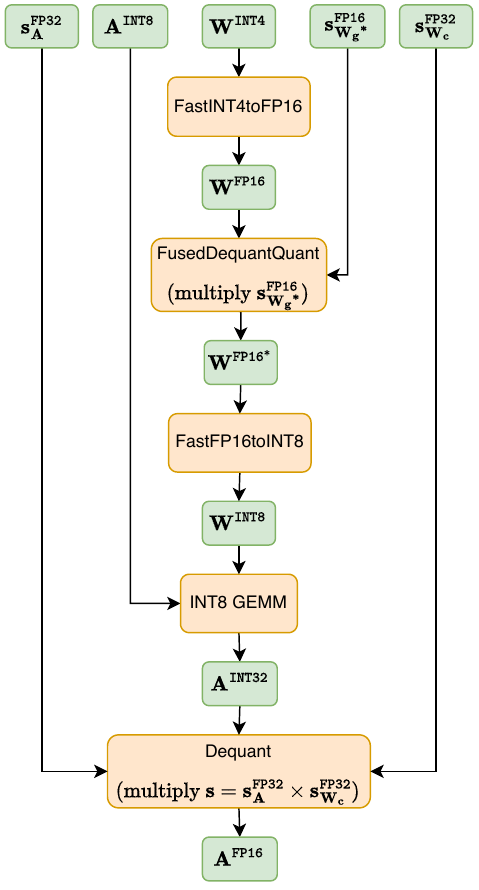}
\caption{The W4A8 GEMM for per-group weight quantization. $\mathbf{s_{W_c}}$ represents the per-channel quantization scale converting from dequantized FP16 weights back to INT8 weights. $\mathbf{s_{W_g}^{*}}$ represents the fused per-group scale converting from INT4 weights to INT8 weights.}
\label{fig:fig4}
\vspace{-0.1in}
\end{figure}

\begin{figure}[ht]
\centering
\includegraphics[scale=0.6]{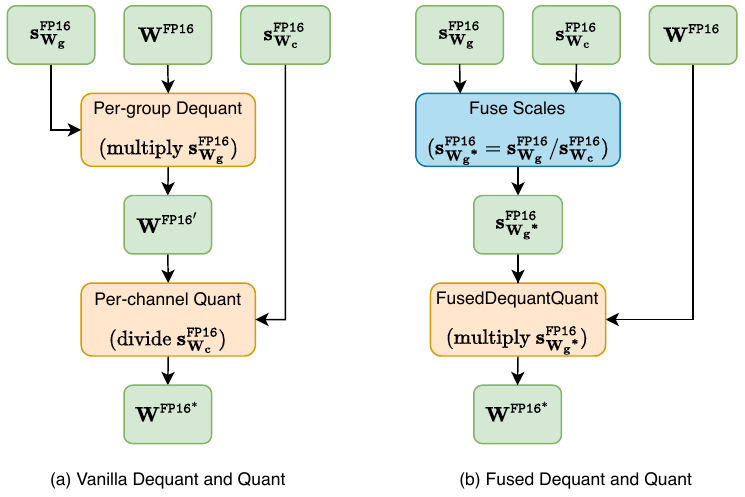}
\caption{Design of \texttt{FusedDequantQuant}. $\mathbf{s_{W_g}}$ represents the per-group quantization scale converting from FP16 weights to INT4 weights.
The orange sections indicate \orange{online operations}, and the blue sections indicate \cyan{offline operations}.}
\label{fig:fig5}
\vspace{-0.1in}
\end{figure}

\subsubsection{Per-group Weight Quantization.}
The W4A8 GEMM tailored for per-group weight quantization is detailed in Figure \ref{fig:fig4}. This approach involves a more complex set of operations prior to the \texttt{INT8 GEMM} compared to the per-channel weight quantization. Although the conversion from INT4 to INT8 weights is still necessary, additional steps are required due to the unique per-group quantization scales that are not directly compatible with standard GEMM procedures.

Initially, INT4 weights are converted to FP16 using the \texttt{FastINT4toFP16} \citep{kim2022says} method. Subsequently, each group's weight is dequantized by multiplying it by its respective scale. To re-quantize into INT8, the result is divided by the per-channel quantization scale. These two steps are elegantly merged into a single operation, \texttt{FusedDequantQuant}, by dividing the per-group scale by the per-channel scale, as illustrated in Figure \ref{fig:fig5}.
The final stage of the process involves converting the FP16 results of \texttt{FusedDequantQuant} back into INT8 format to proceed with the \texttt{INT8 GEMM}. Similar to the per-channel method, the INT32 output from the \texttt{INT8 GEMM} undergoes dequantization to FP16.

\begin{figure}[ht]
\centering
\includegraphics[scale=0.6]{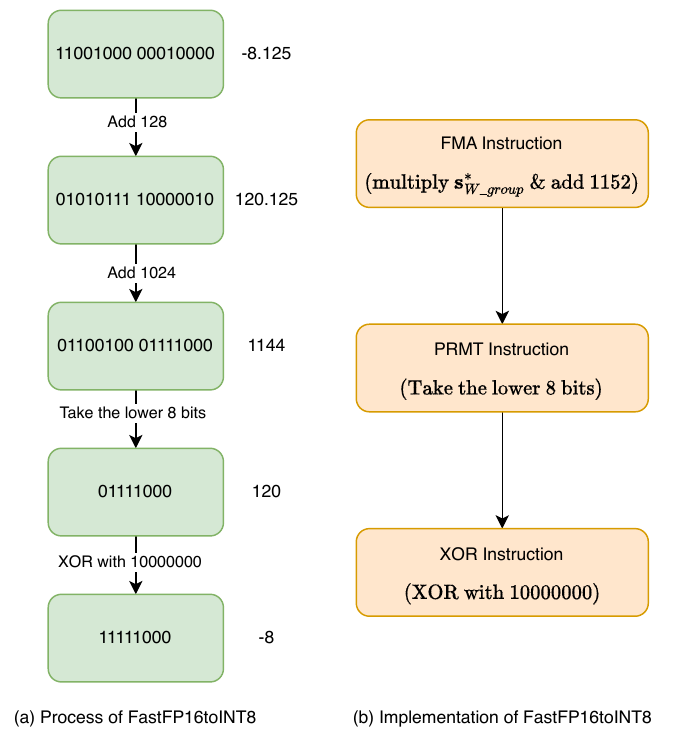}
\caption{Design of \texttt{FastFP16toINT8}.}
\label{fig:fig6}
\vspace{-0.1in}
\end{figure}

The conventional instruction for converting FP16 to INT8 is significantly slow, which hinders the overall GEMM performance. To tackle this issue, we introduce a fast conversion technique, \texttt{FastFP16toINT8}, as depicted in Figure \ref{fig:fig6}. The conversion begins by adjusting the FP16 values, adding $128$ to shift them into the range of $[0.0, 255.0]$, which corresponds to the representational span of UINT8. This is followed by an addition of $1024$, effectively aligning the 8 bits of UINT8 within the lower segment of FP16's mantissa. The final step involves extracting these lower 8 bits from FP16 and applying an XOR operation with $0\textrm{x}80$ to achieve the desired INT8 format. In practice, the fusion of the per-group quantization scale multiplication with the addition of the magic number $128 + 1024$ is executed through a single \texttt{FMA} (Fused Multiply-Add) instruction. Furthermore, the process leverages two expedient bit manipulation instructions, \texttt{PRMT} and \texttt{XOR}, to finalize the conversion.


\begin{table*}[t]
\centering
\begin{small}
\renewcommand{\arraystretch}{1.5}
\begin{tabular}{llccccccc}
\toprule
\multirow{2}*{\#Bits} & \multirow{2}*{Method} & \multicolumn{3}{c}{LLaMA-1} & \multicolumn{3}{c}{LLaMA-2} & LLaMA-3\\
\cline{3-9}
&& 7B & 13B & 30B & 7B & 13B & 70B & 8B\\
\midrule
FP16 & - & 5.68 & 5.09 & 4.10 & 5.47 & 4.88 & 3.32 & 6.14\\
\hline
W8A8 & SmoothQuant & 5.78 & 5.17 & 4.28 & 5.58 & 4.93 & 3.39 & 6.25\\
\hline
\multirow{2}*{W4A16} & GPTQ-g128 & 5.83 & 5.20 & 4.22 & 5.63 & 4.99 & 3.43 & 6.56\\
& AWQ-g128 & 5.78 & 5.19 & 4.21 & 5.60 & 4.97 & 3.41 & 6.54\\
\hline
W4A4 & Atom-g128 & 6.25 & 5.52 & 4.61 & 6.12 & 5.31 & 3.73 & 7.76\\
\hline
\multirow{4}*{W4A8} & QoQ & 5.93 & 5.28 & 4.34 & 5.75 & 5.12 & 3.52 &6.89\\
& QoQ-g128 & 5.89 & 5.25 & 4.28 & 5.70 & 5.08 & 3.47 & 6.76\\
& QQQ & 6.19 & 5.43 & 4.61 & 5.95 & 5.21 & 3.68 & 7.41\\
& QQQ-g128 & 5.87 & 5.24 & 4.30 & 5.71 & 5.01 & 3.50 & 6.64\\
\bottomrule
\end{tabular}
\end{small}
\caption{WikiText2 perplexity with 2048 sequence length. The lower is the better. '-g128' denotes applying per-group quantization on the weights and the group size is 128.
}
\label{tab:tab1}
\end{table*}

\begin{table*}[t]
\centering
\begin{small}
\renewcommand{\arraystretch}{1.5}
\begin{tabular}{lllcccccc}
\toprule
LLaMA-2 & \#Bits & Method & PIQA & ARC-e & ARC-c & HellaSwag & WinoGrande & Avg.\\
\midrule
\multirow{4}*{7B} & FP16 & - & 79.05 & 74.58 & 46.25 & 76.05 & 68.98 & 68.98\\
\cline{2-9}
& W4A4 & Atom-g128 & 75.14 & 52.99 & 38.40 & 69.37 & 62.75 & 59.73\\
\cline{2-9}
& \multirow{4}*{W4A8} & QoQ & 77.64 & 72.81 & 43.60 & 74.00 & 68.03 & 67.22 \\
&& QoQ-g128 & 78.07 & \textbf{73.32} & \textbf{44.80} & \textbf{74.98} & \textbf{68.59} & \textbf{67.95}\\
&& QQQ & 77.42 & 69.15 & 42.15 & 73.54 & 65.98 & 65.65\\
&& QQQ-g128 & \textbf{78.51} & 72.94 & 44.37 & 74.53 & 67.01 & 67.47\\
\hline
\multirow{4}*{13B} & FP16 & - & 80.52 & 77.44 & 49.06 & 79.38 & 72.22 & 71.72\\
\cline{2-9}
& W4A4 & Atom-g128 & 76.50 & 57.49 & 42.32 & 73.84 & 67.40 & 63.51\\
\cline{2-9}
& \multirow{4}*{W4A8} & QoQ & 79.71 & 75.97 & 48.38 & 77.80 & 70.96 & 70.56\\
&& QoQ-g128 & 79.43 & \textbf{77.06} & \textbf{48.81} & 78.35 & 70.48 & 70.83\\
&& QQQ & 79.43 & 74.75 & 48.12 & 77.27 & 70.32 & 69.98\\
&& QQQ-g128 & \textbf{79.98} & 76.64 & 48.55 & \textbf{78.63} & \textbf{71.82} & \textbf{71.13}\\
\bottomrule
\end{tabular}
\end{small}
\caption{Zero-shot accuracy on five common sense tasks with 2048 sequence length. The higher is the better.
}
\label{tab:tab2}
\end{table*}

\section{Experiments}\label{section:sec5}
\subsection{Setups}
In this section, we introduce the experimental configurations in models, datasets, algorithm and inference system.

\textbf{Models and Datasets.} We conduct the experiments using the LLaMA series \cite{touvron2023llama, touvron2023llama2}. We evaluate the model performance on language modeling and zero-shot tasks. For the language modeling aspect, we utilize the WikiText2 \citep{merity2016pointer} dataset to measure perplexity. For the zero-shot tasks, we extend the evaluation to a variety of datasets, including PIQA \citep{bisk2020piqa}, HellaSwag \citep{zellers2019hellaswag}, WinoGrande \citep{sakaguchi2021winogrande}, and ARC \citep{clark2018think}, utilizing the \texttt{lm\_eval} \citep{eval-harness} toolkit for a comprehensive analysis. Additionally, we select a subset of 128 sequences from the PILE \citep{gao2020pile} dataset to facilitate calibration.

\textbf{Algorithm.} QQQ is specifically designed to focus on per-token quantization for activations and supports both per-channel and per-group quantization for weights. To optimize inference speed, QQQ exclusively uses symmetric quantization for both activations and weights, thus avoiding the additional computational overhead associated with asymmetric quantization.

\textbf{Inference System.} The inference system that we have develop to support QQQ is based on vLLM\footnote{\url{https://github.com/vllm-project/vllm}} \citep{kwon2023efficient} v0.4.1. The system's capabilities are further enhanced by the implementation of W4A8 GEMM kernels, which are based on the advanced Marlin\footnote{\url{https://github.com/IST-DASLab/marlin}} \citep{frantar2024marlin} kernels. All experimental procedures are conducted on NVIDIA A100 80G GPUs under PyTorch 2.1.2 with CUDA 11.8.

\subsection{Model Performance}
We perform a comprehensive comparison of QQQ method against leading LLM quantization methods, including SmoothQuant \citep{xiao2023smoothquant}, GPTQ \citep{frantar2022gptq}, AWQ \citep{lin2023awq}, Atom \citep{zhao2023atom}, and the recently introduced W4A8 method QoQ \citep{lin2024qserve}. By default, weights employ per-channel quantization. GPTQ, AWQ, and QoQ utilize asymmetric quantization for weights, whereas SmoothQuant, Atom, and QQQ opt for symmetric quantization. In terms of activation quantization, Atom stands out by implementing per-group quantization, while the remaining methods, including QQQ, employ per-token quantization. Across all methods under consideration, activations apply symmetric quantization.

Table \ref{tab:tab1} presents a detailed perplexity comparison on the WikiText2 dataset. The per-group QQQ demonstrates competitive performance with with W8A8 SmoothQuant, W4A16 GPTQ, and W4A16 AWQ across various models. For example, per-group QQQ only increases perplexity by up to 0.13 on LLaMA-2-13B compared with them. When benchmarked against W4A4 Atom, both the per-channel and per-group QQQ consistently outperform Atom on all tested models. For instance, per-group QQQ achieves a significant reduction in perplexity by 0.41 on LLaMA-2-13B when compared to Atom. In comparison with W4A8 QoQ, per-group QQQ outperforms per-group QoQ on the majority of models, exemplified by a 0.12 decrease in perplexity on LLaMA-3-8B. However, it is observed that per-channel QQQ falls slightly behind per-channel QoQ. This divergence is hypothesized to be due to QoQ's application of asymmetric quantization on weights, as opposed to the symmetric quantization employed by QQQ.

The zero-shot accuracy evaluation across five common sense tasks is showed in Table \ref{tab:tab2}. The QQQ method exhibits a significant performance edge over Atom across all tasks and model sizes. For instance, the per-group QQQ achieves a 4.79\% higher accuracy than Atom on the HellaSwag task when utilizing the LLaMA-2-13B model. In comparison with QoQ, QQQ demonstrates comparable results, especially when per-group quantization is applied. For instance, QQQ shows a marginal improvement of 0.3\% over QoQ with the LLaMA-2-13B model on average.

\begin{figure*}[ht]
\centering
\includegraphics[scale=0.53]{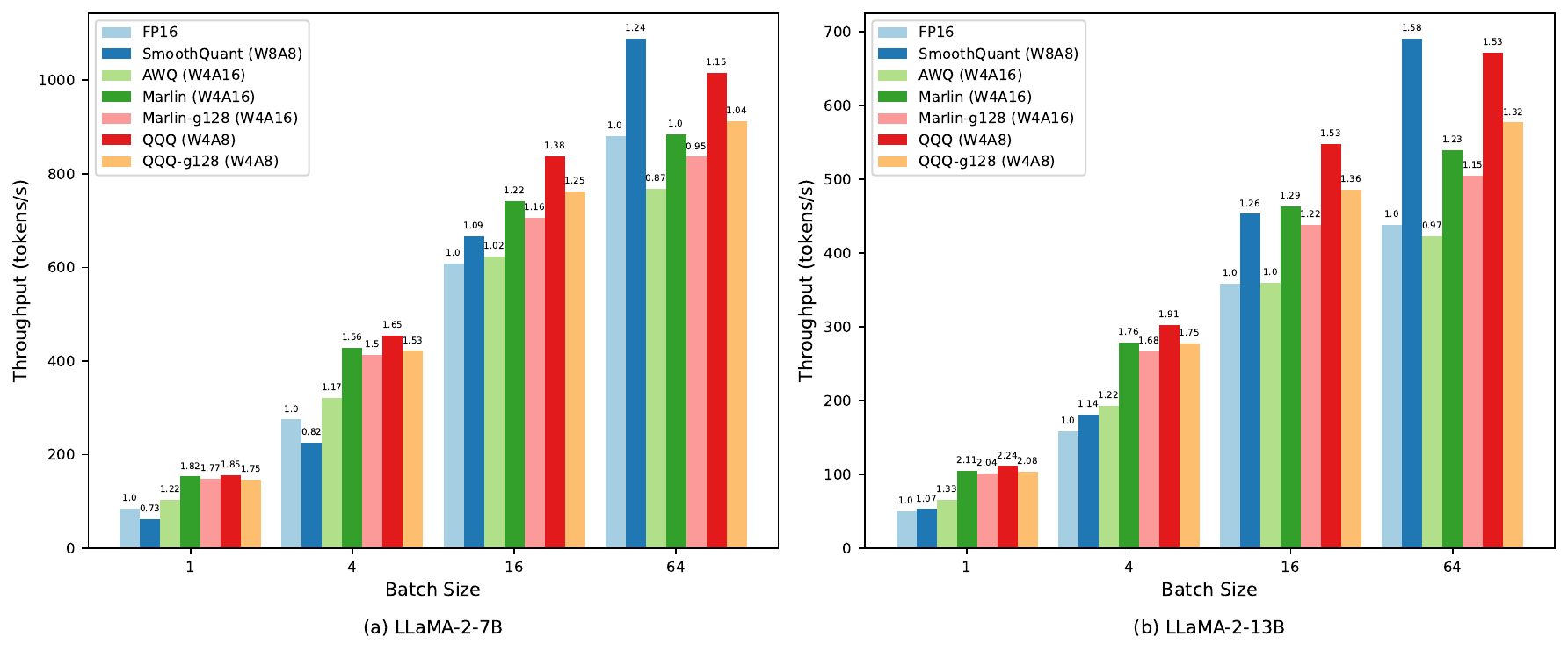}
\caption{Same-batch throughput comparison of quantized LLaMA-2 models under various batch size.}
\label{fig:fig7}
\vspace{-0.1in}
\end{figure*}

\subsection{Speedup}
In the evaluation of inference efficiency, QQQ is benchmarked against a suite of quantization methods including SmoothQuant\footnote{\url{https://github.com/vllm-project/vllm/pull/1508}} \citep{xiao2023smoothquant}, AWQ \citep{lin2023awq}, and GPTQ \citep{frantar2022gptq}, all of which are integrated within the vLLM \citep{kwon2023efficient}. For optimal inference speed, we leverage the Marlin \citep{frantar2024marlin} implementation of GPTQ. All methods are tested under conditions that enable continuous batching and paged attention. The benchmarks are conducted with an input sequence length of 1024 and an output sequence length of 128.

The throughput comparison outlined in Figure \ref{fig:fig7} indicates that QQQ outperforms the alternative methods across a range of batch sizes in most cases. Specifically, QQQ attains a substantial speed increase, reaching up to 2.24$\times$, 2.10$\times$, 1.59$\times$, and 1.25$\times$ faster than FP16, SmoothQuant, AWQ, and Marlin on the LLaMA-2-13B model, respectively. When contrasted with W4A16 AWQ and W4A16 Marlin, QQQ displays a comparable inference speed at smaller batch sizes but gains a significant advantage as the batch size increases. Against W8A8 SmoothQuant, QQQ stands out with an approximate 2.0$\times$ speedup for smaller batches and maintains a similar level of acceleration even at a batch size of 64. Moreover, QQQ's performance enhancement is more pronounced on larger models, as evidenced by the greater speedup achieved on LLaMA-2-13B compared to LLaMA-2-7B.

\begin{figure}[ht]
\centering
\includegraphics[scale=0.50]{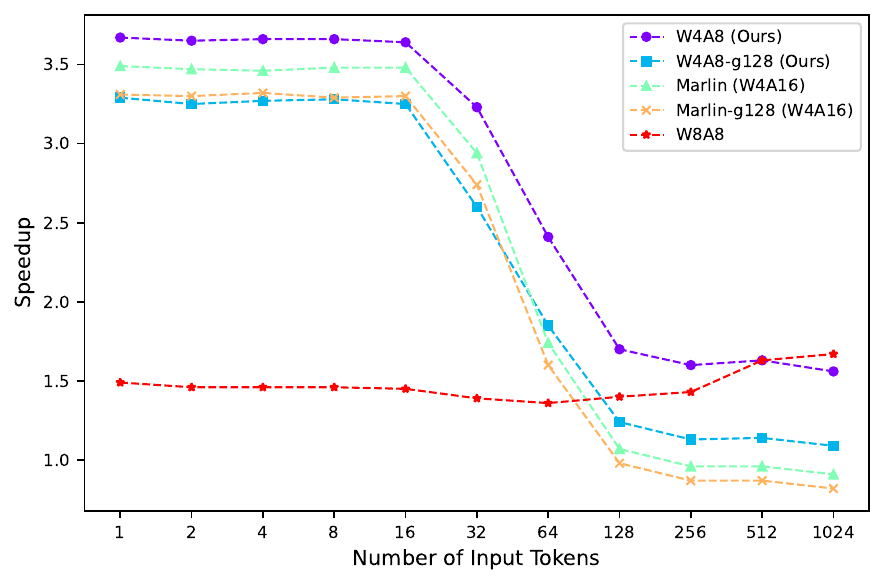}
\caption{Speedup over PyTorch FP16 GEMM (Calling CUTLASS) of all GEMMs under different numbers of input tokens. The weight matrix size is ($N=8192, K=21760$).}
\label{fig:fig8}
\vspace{-0.1in}
\end{figure}

\subsection{W4A8 GEMM Studies} 
We conduct a series of experiments to demonstrate the capabilities of our W4A8 GEMMs. By keeping the weight matrix size constant and varying the number of input tokens, we evaluate the speed improvements over the PyTorch FP16 GEMM benchmark. The results are depicted in Figure \ref{fig:fig8}.

Our W4A8 GEMMs demonstrate impressive speedups, achieving up to 3.67$\times$, 2.51$\times$, and 1.71$\times$ faster performance compared to PyTorch FP16 GEMM, W8A8 GEMM, and W4A16 Marlin GEMM, respectively. Specifically, W4A8 GEMMs show a significant advantage when the token count is below 64 compared to W8A8 GEMM. With an increasing number of tokens, the per-group W4A8 GEMM maintains a similar level of efficiency, whereas the per-channel W4A8 GEMM tends to be slower. This discrepancy is attributed to the higher type conversion overhead in per-group W4A8 GEMM.
Compared to Marlin GEMM, our GEMMs demonstrate superior performance under the majority of token counts, especially per-group GEMM.
As the token count rises, Marlin GEMM's performance gradually declines, falling behind even the FP16 GEMM, while our GEMMs consistently outperform the FP16 benchmark. This clearly indicates the superior efficiency of INT8 Tensor Cores over FP16 Tensor Cores in compute-bound scenarios.

\subsection{Ablation Studies}

\begin{table}[t]
\centering
\begin{small}
\renewcommand{\arraystretch}{1.3}
\setlength{\tabcolsep}{0.8mm}{
\begin{tabular}{llccc}
\toprule
Granularity & Model & B & B+AS & B+AS+GPTQ\\
\midrule
\multirow{7}*{Per-channel} & LLaMA-1-7B & 6.93 & 6.77 & \textbf{6.19} \\
\cline{2-5}
& LLaMA-1-13B & 5.84 & 5.84 & \textbf{5.43}\\
\cline{2-5}
& LLaMA-1-30B & 5.04 & 4.89 & \textbf{4.61}\\
\cline{2-5}
& LLaMA-2-7B & 7.36 & 7.29 & \textbf{5.95}\\
\cline{2-5}
& LLaMA-2-13B & 5.47 & 5.43 & \textbf{5.21}\\
\cline{2-5}
& LLaMA-2-70B & 4.07 & 3.88 & \textbf{3.68}\\
\cline{2-5}
& LLaMA-3-8B & 10.88 & 9.46 & \textbf{7.41}\\
\hline
\multirow{7}*{Per-group} & LLaMA-1-7B & 6.16 & 5.96 & \textbf{5.87}\\
\cline{2-5}
& LLaMA-1-13B & 5.43 & 5.32 & \textbf{5.24}\\
\cline{2-5}
& LLaMA-1-30B & 4.51 & 4.34 & \textbf{4.30}\\
\cline{2-5}
& LLaMA-2-7B & 5.86 & 5.77 & \textbf{5.71}\\
\cline{2-5}
& LLaMA-2-13B & 5.10 & 5.06 & \textbf{5.01}\\
\cline{2-5}
& LLaMA-2-70B & 3.67 & 3.52 & \textbf{3.50}\\
\cline{2-5}
& LLaMA-3-8B & 7.11 & 6.96 & \textbf{6.64}\\
\bottomrule
\end{tabular}
}
\end{small}
\caption{WikiText2 perplexity of LLaMA series when applying various quantization techniques. 'B' represents the Baseline, which refers to the vanilla W4A8 method without any additional quantization techniques. 'AS' denotes Adaptive Smoothing.
}
\label{tab:tab3}
\end{table}

\subsubsection{Impact of Quantization Techniques}
Here we examine the impact of different quantization techniques on model performance. We use the standard W4A8 method, which lacks any specialized quantization techniques, as the baseline for comparison. The perplexity results for the LLaMA series are summarized in Table \ref{tab:tab3}. Overall, the combination of adaptive smoothing with GPTQ proves to be the most effective method, yielding the lowest perplexity scores. For instance, this approach reduces perplexity by 0.47 in per-group quantization compared with the baseline.

\begin{figure}[ht]
\centering
\includegraphics[scale=0.5]{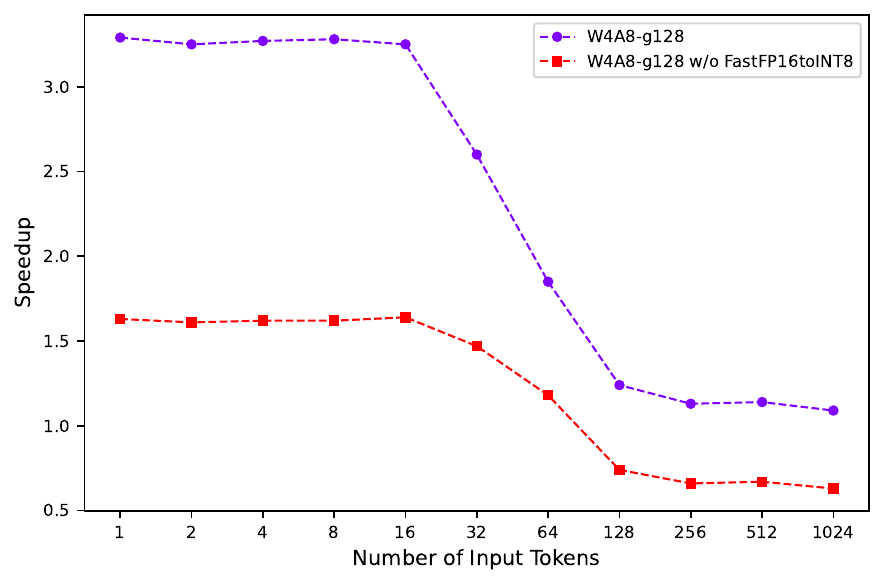}
\caption{Speedup over PyTorch FP16 GEMM (Calling CUTLASS) of two per-group W4A8 GEMMs under different numbers of input tokens. The weight matrix size is ($N=8192, K=21760$).}
\label{fig:fig9}
\vspace{-0.15in}
\end{figure}
\subsubsection{Impact of FastFP16toINT8}
Here we explore the performance enhancements achieved in per-group W4A8 GEMM through the implementation of the \texttt{FastFP16toINT8} conversion technique. We benchmark the optimized W4A8 GEMM against the conventional W4A8 GEMM, which utilizes the standard FP16 to INT8 conversion instruction. As illustrated in Figure \ref{fig:fig9}, the GEMM utilizing \texttt{FastFP16toINT8} demonstrates a significant increase in speedup compared to its counterpart without this optimization, across a range of input tokens. The \texttt{FastFP16toINT8} technique can accelerate per-group W4A8 GEMM by up to 2.02$\times$. This finding confirms that the FP16 to INT8 conversion process is a critical bottleneck, and our \texttt{FastFP16toINT8} effectively enhances the efficiency of per-group W4A8 GEMM by mitigating this constraint.

\section{Conclusion}\label{section:sec6}
In this paper, we propose a quality W4A8 quantization method optimized for hardware efficiency. QQQ leverages the power of adaptive smoothing combined with Hessian-based quantization compensation to significantly improve the performance of quantized models. In addition, we meticulously develop W4A8 GEMM kernels to maximize the inference speed. Through extensive experiments, QQQ demonstrates superior performance over existing state-of-the-art LLM quantization methods in terms of both performance and speed. We believe QQQ can serve as a powerful tool that opens up access to the deployment and application of LLMs.

\section*{Limitations}
QQQ method has two limitations: 1) The two-stage quantization method requires more time for quantizing models. 2) The mixed-precision GEMMs only support 4-bit weights for now.

\bibliography{acl}
\end{document}